\documentclass[letterpaper]{article}

  \usepackage{times}
  \usepackage{helvet}
  \usepackage{courier}
  \usepackage{url}
  \usepackage{graphicx}
\usepackage{aaai18}
\usepackage[linesnumbered]{algorithm2e}
\usepackage{times}
\usepackage{amsmath,enumerate,amsfonts,amssymb,amsthm}
\usepackage{mathrsfs}
\usepackage{stmaryrd}
\usepackage{multicol} 
\usepackage{multirow}
\usepackage{xfrac}
\usepackage{threeparttable}
\usepackage{pgf}
\usepackage{tikz}
\usetikzlibrary{arrows,automata}
\usepackage[latin1]{inputenc}
\usepackage{verbatim}
\usepackage{graphicx}
\usepackage{caption}
\usepackage{subcaption}
 \usepackage{pifont}
 \usepackage{diagbox}
 \usepackage{enumitem}
\newtheorem{defn}{Definition}

\newtheorem{ex}{Example}

\newcommand{\freddy}[1]{{\color{black}#1}}

\newcommand{\mlmodel}[1]{\mathbb{M}_{#1}}

\newcommand{\ont}{\mathcal{O}}
\newcommand{\cp}[1]{\mathbb{C}_{#1}}

\newcommand{\repi}{\mathfrak{I}}

\newcommand{\matrixcps}[1]{\mathcal{M}_{#1}}
\newcommand{\prj}[2]{\mathcal{P}_{#1}(#2)}
\newcommand{\setcps}[1]{\mathcal{C}_{#1}}
\newcommand{\dist}[2]{\texttt{dist}(#1,#2)}

\newcommand{\semuplift}{\mathfrak{F}_s}
\newcommand{\repsel}{\mathfrak{F}_r}

\newcommand{\conceptmapping}{\mathfrak{F}_{io}}

\newcommand{\score}[1]{s_{#1}}

\DeclareMathAlphabet{\mathpzc}{OT1}{pzc}{m}{it}
\DeclareMathAlphabet\mathbfcal{OMS}{cmsy}{b}{n}
\DeclareMathOperator*{\argmax}{\arg\!\max}

\title{Semantic Explanations of Predictions}

\author{
Freddy L\'{e}cu\'{e}\thanks{Authors contributed equally to the work.}\\
INRIA, Sophia Antipolis, France \\
Accenture Labs, Dublin, Ireland\\
freddy.lecue@inria.fr \\ 
\And 
Jiewen Wu$^\ast$\\
A*STAR Artificial Intelligence Initiative\\Institute for InfoComm Research\\Singapore\\
jiewen.wu@acm.org
}
\begin{document}
\maketitle
\begin{abstract}
%
The main objective of explanations is to transmit knowledge to humans. This work proposes to  construct \emph{informative} explanations for predictions made from machine learning models. Motivated by the observations from social sciences, our approach selects data points from the training sample that exhibit special characteristics crucial for explanation, for instance, ones contrastive to the classification prediction and ones representative of the models. Subsequently, semantic concepts are derived from the selected data points through the use of domain ontologies. These concepts are filtered and ranked to produce informative explanations that improves human understanding. The main features of our approach are that (1) knowledge about explanations is captured in the form of ontological concepts, (2) explanations include contrastive evidences in addition to normal evidences, and 
(3) explanations are user relevant.
%
%
\end{abstract} 

\section{Introduction}

%
Machine learning, particularly deep learning, has attracted attentions from both industries and academia  over the years. The algorithmic advancement has spurred near-human level accuracy applications such as neural machine translation \cite{googleMT}, novel methods including Generative Adversarial Networks \cite{DBLP:conf/nips/SalimansGZCRCC16} and Deep Reinforcement Learning \cite{mnih-dqn-2015}, among other things.
Although highly scalable, accurate and efficient, most, if not all, of the machine learning models have exhibited limited interpretability \cite{DBLP:conf/kdd/LouCG12}, which implies humans can hardly explain the final predictions \cite{shmueli2010explain}. The lack of meaningful explanations of prediction would become more problematic when the models are deployed in financial, medical, and public safety domains, among many others. Explanations are indispensable for building the \emph{trust} relationship between human decision makers and intelligent systems making predictions. For instance, both the context and the \freddy{rationale} of any prediction result in medical diagnosis \cite{DBLP:conf/kdd/CaruanaLGKSE15} need to be understood as some of its consequences may be disastrous. In addition to trust,  business owners can demand explanations for more informed decision making and developers can leverage explanations to debugging and maintenance. More stringent requirements have been dictated from legislation to safeguard fair and ethical decision making in general, notably the European Union General Data Protection Regulation  (GDPR) warranting users the ``right to explanation" in algorithmic decision-making \cite{DBLP:journals/corr/GoodmanF16}.

Although there has been a lack of consensus on the definition of \emph{explanations}, we have witnessed multiple avenues of research.  As argued in \cite{mythos}, these efforts generally fall into two (not necessarily disjoint) categories, i.e., one that aims at improving \emph{transparency} of decision making by unveiling the internal mechanism of machine learning models and the other being \emph{post hoc} explanations that justify the predictions generated by the models.  The first category is sometimes referred to as \emph{interpretability}.
Our paper posits itself in the post hoc explanation category, in response to this specific question: 
\begin{equation}\label{q:main}
\emph{Why the input $x$ was labeled $y$?}
\end{equation}
It is obvious that answers to such a question can be subjective. This paper, instead of deriving a complete solution to ``correct" explanations, addresses the issue of \emph{informativeness}  \cite{mythos}. Towards informative explanations, we   investigate certain salient properties on elucidating predictions to human users.  In particular, we observe the following survey findings in social sciences \cite{millerXAI} towards explanations. 
\begin{itemize}[leftmargin=*,labelindent=1em,labelsep=1em]
\item[$O_1$]  Human explanations imply social interaction \cite{conversationalExp}. The implication is that, for machine-generated explanations, it is indispensable to associate \emph{semantic} information  with an explanation (or the elements therein) for effective knowledge transmission to users.
\item[$O_2$] Users  favor contrastive explanations  for understanding of causes \cite{contrastEQ,contrastLipton}. That is, \eqref{q:main}  often implies the question: 
\begin{equation}\label{q:contrast}
\emph{Why the input $x$ was labeled $y$ \emph{instead of $y'$?}}
\end{equation}
\item[$O_3$] Users \emph{select} explanations. Due to the large space of possible explanations and a specific user's understanding of the context, she selects the explanations based on what she believes to be the most \emph{relevant} to her, rather than the most direct or probable causes \cite{conversationalExp}. 
The subjectivity of human choices implies that informative explanations may need to consider personalisation or  contextualisation.
\end{itemize}

This paper proposes a method that leverages semantic concepts drawn from data instances to characterize the aforementioned three observations for explanations, thus enabling more effective human understanding of predictions.

Most of existing approaches focus on data-driven explanation and lack semantic interpretation, which defeats the objective of human-centric explanations.
Instead, our proposed approach exploits the semantics of representative data points in the training samples. 
It works by (i) selecting representative data points and elaborating the decision boundary of classifiers, (ii) extracting and encoding the semantics of such data points using domain ontologies, and (iii) computing informative explanations based on optimizing certain criteria learned from humans' daily explanations.
%

The remainder of the paper subsequently reviews the basics and introduces the problem.
Then, we describe how representative data points are extracted and show how semantics of data point is exploited to derive explanations. Conclusions and future research directions are given in the end.

\section{Related Work}

Interpreting models or predictions dates back to at least twenty years ago. The resurgence of neural nets also attracted a lot of recent research into the area of interpretability of such deep models. To show the position of our proposed approach, we discuss a few representative work in the field of machine learning, from the angles outlined in $O_1$, $O_2$, and $O_3$. 

%
Decision trees and random forests have been studied to extract various levels of model interpretation \cite{treeNN,DBLP:conf/iri/PalczewskaPRN13} together with some degrees of prediction explanation \cite{pang2006pathway}. Although their explanations are tuned to complex stochastic and uncertain rules they naturally expose high visibility on the decision process. 
%
%
\cite{DBLP:conf/icdm/WangRDLKM16} exploit the characteristics of classification models by exploiting and relaxing their decision boundary to approximate explanations.
However explanations remain handcrafted from features and raw data, often as rules which remain very difficult to be generalized.
\cite{DBLP:conf/naacl/LiCHJ16} targeted neural networks and observed the effects of erasing various parts of the data and its features on the model to derive a minimum but representative set, qualified as explanation.
\cite{DBLP:conf/emnlp/LeiBJ16} study similar models and aim at identifying candidate \freddy{rationales} i.e., core elements of the model which aims at generalizing any prediction. 
Instead, \cite{DBLP:conf/nips/KimSD15} focused on placing interpretability criteria directly into the model to ensure fined-grained exploration and generation of explanations.
\cite{kddexplain} elaborated a model-agnostic technique. To this end any test data is re-sampled and approximated using training data, which is then used as a view, or explanation of the predictions and model. Note that our work here is not model-agnostic and focuses more on the semantic interpretation to achieve higher level of  informativeness. 
Leveraging \emph{contrastive} information for explaining the predictions has seen its applications in image classification, e.g., \cite{Vedantam_2017_CVPR}, which justifies why an image describes a particular, fine-grained concept as opposed to the distractor concept.

Towards human-centric explanations, \cite{recsysdesign} designed some general properties of effective explanations in recommender systems.  \cite{ijcai17justification} focuses on combining instance-level and feature-level information to provide a framework that generalizes several types of explanations.
A more complete survey on human-centric explanations is also available in \cite{millerXAI}, highlighting research findings from social sciences.


%

\section{Problem Statement}\label{sec:Background}
We focus on the predictions given by (w.l.o.g., binary) classifiers, where data points are partitioned into sets, each of which belonging to one class. The partition surface is a \emph{decision boundary}. Before delving into the technical details, we first define the problem to be addressed.
%

%
%
 
%
\subsection{Ontology}

%
%
%
An ontology $\ont$ describes the concept hierarchy of domain knowledge. A concept, denoted by $\cp{}$, represents a type of objects. 
%
The most common relationship between concepts is subsumption (is-a), denoted $\sqsubseteq$. For instance, Human $\sqsubseteq$ Animal w.r.t. some ontology. 
An ontology can define many different types of semantic relationships beyond just is-a relationship, e.g., hasChild, hasParent, and so on. The hierarchical relations of concepts in $\ont$ can be described as a graph for easy manipulation: each concept is a vertex, while the semantic relationship is a directed edge. An edge may be weighted to indicate how strong the semantic relation is between the concepts.

%
%

%




  
\subsection{Explanation Problem Statement}

An informative explanation, the objective of this paper, can be defined based on observations $O_1$-$O_3$. 
Without loss of generality, consider a binary classifier, $\mlmodel{}$, and a prediction $y_i$ of some given data point $x_i$,  intuitively, the aim is to find a set of \emph{human-understandable descriptions}  of $y_i$ with respect to $\mlmodel{}$ and $x_i$. 
To ease the presentation, a classifier is abused as a function, too. That is, $\mlmodel{}(x_i) = y_i$ means that $x_i$ is predicted to be of label $y_i$ by the classifier $\mlmodel{}$.
A formal definition now follows. 
\begin{defn}{\bf (Informative Explanation)}{\label{defn:explanation}}\\
Let $\mlmodel{}$ be a binary classifier, $X$ be the set of training data points, $x_i$ be a test data point with $y_i$=$\mlmodel{}(x_i)$, and $\setcps{}$ be a set of concepts. 
We define a data point selection function $\repsel$:$X\rightarrow \{0,1\}$ and a semantic uplift function $\semuplift:2^X\rightarrow {2^\setcps{}}$.

An informative explanation is $e=\setcps{}^{+}\cup\setcps{}^{-}$, where $\setcps{}^{+}=\semuplift(\{x_j~\mid~ j\ne i, \repsel(x_j)=1, \text{ and }\mlmodel{}(x_j) = y_i\})$  and $\setcps{}^{-}=\semuplift(\{x_k~\mid~ k\ne i, \repsel(x_k)=1, \text{ and }\mlmodel{}(x_k)\ne y_i\})$ . 
%
%
\end{defn}
Observe that the definition of the semantic uplift function, $\semuplift$, implies that a set of data points can be assigned multiple ontological concepts.

\noindent\textbf{Algorithmic considerations}
In addition to defining functions $\repsel$ and $\semuplift$, two more conditions are imposed on the algorithm design: (1) $e$ must be concise to meet observations $O_1$ and $O_2$. So, the size of $e$ needs to optimized using quantifiable measures. (2) the content of $e$  needs to show contrastive information and ranking is necessary to allow for user choices based on relevancy, as discussed in $O_3$. 
%
%

%
\noindent\textbf{Assumption} 
To semantically interpret data points, the raw feature must be (at least partially) semantically meaningful, so that semantics is available from the beginning.  In practice most datasets have textual descriptions, and, in the rare cases where the raw features lack any descriptions, advices from dataset owners or domain experts can be sought. 
Note that the proposed approach assumes nominal features are expressed in one-hot encoding.


%
%
\freddy{
\begin{ex}{\bf (Explanation of Classification Prediction)}\\
The rest of the paper uses the dataset Haberman's Survival from the UCI repository\footnote{http://archive.ics.uci.edu/ml/datasets/} as an running example. 
The dataset contains cases from a study conducted 1958-1970 at the University of Chicago's Billings Hospital on the survival of patients who had surgery for breast cancer.
The task aims at classifying patients into those (1) survived 5 years or longer or (2) died within 5 year using the predictors: age, year of operation, and positive axillary nodes detected. 
We aim at identifying the informative explanations, as in Definition \ref{defn:explanation}, for any predicted data point w.r.t a classifier $\mlmodel{}$, the $306$ training data points and a domain ontology $\ont$.
\end{ex}
}

\subsection{Organization}
The organization of the remainder is as follows. We first describe the data point selection function $\repsel$, which chooses the most interesting data points for the defined problem. From the chosen data points, we then show how concepts can be drawn, enhanced by consulting ontologies, reduced for succinctness, and finally ranked for user choices. 

\section{Identifying Representative Training Data}\label{sec:DB}

To generate an informative explanation, we need to first consider how to find representative data points, i.e., the function $\repsel$.
For the sake of clarity, we concentrate on binary classification (positive or negative) problems and two representative machine learning models, one is a linear classifier using {\bf L}ogistic {\bf R}egression ({\bf LR}), while the other is the non-linear classifier using {\bf k}-{\bf N}earest {\bf N}eighbour ({\bf k-NN}).
The approach can be easily extended to multi-class classifiers. 
 

%
\subsection{Decision Boundary of Classification Models}\label{sec:DBCM}


%
We now review how to compute the decision boundaries of the two models.

\vspace{0.1cm}
\noindent {\bf $\bullet$ {LR Models}} are captured as follows:
\begin{align}{\label{eq:LRM}}
p = \frac{1}{1+e^{-Y}}, \text{ where } Y = \sum_{i=0}^{n}w_i\cdot X_i, 
\end{align}  
where $p$ denotes the probability of being in the positive class. $w_i$ are the parameters for the $n$ given predictors $X_i$. In particular, $X_0$ is the constant $1$ and $w_0$ is the intercept. 

\vspace{0.1cm}
\noindent {\bf $\bullet$ {LR Decision Boundary}} is computed using \eqref{eq:DBLR} considering the positive class with $p\ge 0.5$ in \eqref{eq:LRM}.
\begin{align}\label{eq:DBLR}
\sum_{i=0}^{n}w_i\cdot X_i = 0. 
\end{align}  

\noindent {\bf $\bullet$ {k-Nearest Neighbour (kNN) Models}} are captured as:
\begin{align}
Y = \frac{1}{k}\cdot\sum_{x_i\in n(x)} y_i, 
\end{align}  
where $n(x)$ is the neighbourhood of $x$ defined by the $k$ closest data points $x_i$ in the training sample and $y_i$ are the respective responses of $x_i$.

\vspace{0.1cm}
\noindent {\bf $\bullet$ {k-NN Decision Boundary}} for the data points belonging to the positive / negative class is computed by elaborating the convex hull \cite{chazelle1993optimal} of all points in the respective class. 
%
%
Given datasets of $n$ points in $d$ dimensions, the convex hull, also known as the smallest convex envelope that contains all $n$ points, can be efficiently computed in $O(n\log n)$ for $d\in\{2, 3\}$. However, in the case of high dimensions, the worse case complexity becomes $O(n^{\lfloor{d/2}\rfloor})$. Moreover, the convex hulls, represented by the points, have an average size of $O(n\log^{d-1}n)$ \cite{convexavg}.
In practice, it is reasonable to obtain an approximation of the convex hulls for high dimensional datasets. For instance, the algorithm proposed in \cite{convexapp}, of which the time complexity is insensitive to $d$ and the size of an approximate convex hull is user-specified.

We will use a single dataset throughout this paper as an example to illustrate how the proposed approach is applied to a classifier to obtain explanations. 

\begin{ex}{\bf (k-NN Decision Boundary)}\\
\freddy{Among the $306$ data points of Haberman's Survival dataset, a random point is chosen to be predicted, while the remaining $305$ are considered to be training sample.}
The convex hull, \freddy{as decision boundary of the model}, consists of 42 points, as blue triangle marks in Figure~\ref{fig:ch}. 
The convex hull, represented by blue triangle points, is computed all the training data points in Haberman's Survival dataset. The red cross-squared point in the lower left is the test data point, and the rest of all training data points, the yellow rounded ones, are enclosed by the convex hull. 

\begin{figure}[ht]
\begin{center}
\includegraphics[width=0.22\textwidth]{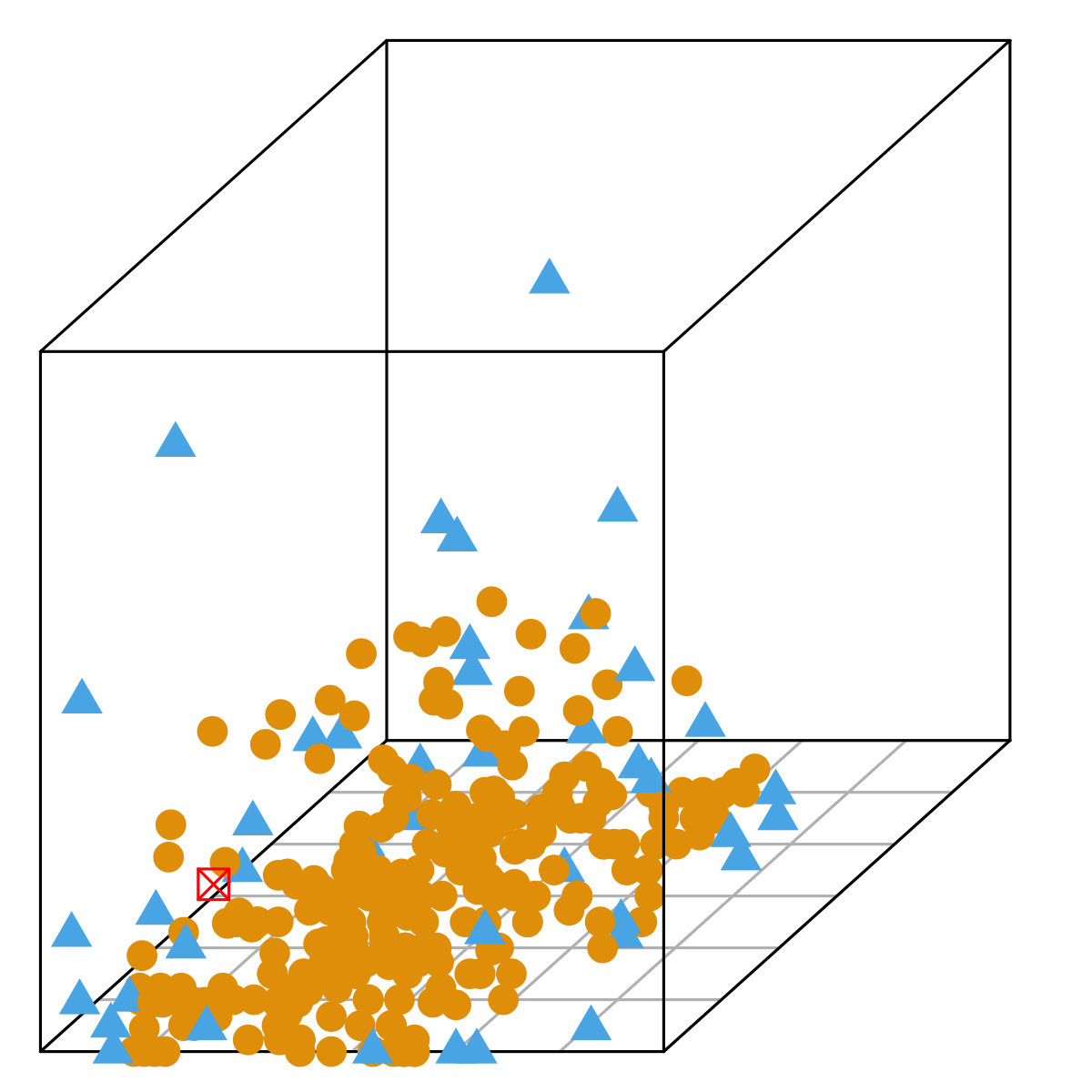}
\end{center}
\vspace{-0.2cm}
\caption{k-NN Decision Boundary. }
\label{fig:ch}
\end{figure}
\end{ex}
  
%
%

%
\vspace{-0.4cm}
\subsection{Representative Data via Decision Boundary} \label{sec:CDPDB}

We illustrate our approach on identifying representative data {points} via the decision boundary.
%
%
%
%
The set of \emph{representative} data points $\repi$ is computed from the decision boundary of models, reflecting the \emph{extreme} cases, and from  the neighbours of the test data points, reflecting the \emph{local} context. Technically, we define $\repi=\repi_g\cup\repi_l$, and we compute the two subsets separately to obtain $\repi$. To distinguish between the different classifiers, these set notations will use superscripts accordingly, e.g., $\repi^{LR}$.
By combining both types of data points, we aim at identifying elements for explanations that meets the objectives $O_1$-$O_3$. 

%

%
\vspace{0.1cm}
\noindent {\bf $\bullet$ LR Representative Data Points} 
Consider a test data point $x_0$  predicted to be of class  $y_0$. 
$\repi_g$ is constructed by selecting data points that have a (standard Euclidean) distance to the decision boundary i.e., a line, within certain threshold $t_g$.  We also consider the proximity between the data points and the test data point $x_0$. That is, neighboring points of $x_0$ will be included, denoted $\repi_l$. The distance between a neighbor and $x_0$ should be within the threshold $t_l$.

The set $\repi_g$ is obtained as follows. First all instances on or close to the decision boundaries are potential elements in $\repi_g$. There is a tradeoff between the size of $\repi_g$ and the representativeness of $\repi_g$. To solve this problem, we find $X_v$ such that $X_v$ has the largest variance among all features (or, the most important feature can be chosen if feature importance is available from the model).  
First all data points that are close enough (determined by the threshold $t_g$) to the decision boundary are collected. 
In the context of LR, we have:  
\begin{align}
\repi_{g}^{LR} &= \{x_g ~\mid~ \frac{ \lvert\sum_{i=0}^{n} w_{i}\cdot X_{ig} \rvert }{ \sqrt{\sum_{i=1}^{n} w_{i} } } \le t_g  ~\wedge~ y_g=y_0  \}\label{DBLR1}\\
\label{locallr}\repi_{l}^{LR} &= \{x_j~\mid~d(x_j, x_0)\le t_l ~\wedge~ y_j=y_0\}
\end{align}  

Note that  the class labels of $x_j$ must be the same as that of the input $x_0$. Finally, we can obtain the set of representative points by combining the two sets: $\repi^{LR} = \repi_{g}^{LR} \cup \repi_{l}^{LR}$. It also follows that, for an arbitrary data point $x'$, $\repsel(x')=1$ if $x'\in\repi^{LR}$, and $\repsel(x')=0$ otherwise. The definition of $\repsel$ in the k-NN case is the same, and is thus omitted.

\vspace{0.1cm}
\noindent {\bf $\bullet$ k-NN Representative Data Points}. The local neighbors in k-NN, represented as $\repi_{l}^{kNN}$, are computed in the same manner as in \eqref{locallr}. A simple version of $\repi_{g}^{kNN}$ is defined as follows: 
\begin{align}
\repi_{g}^{kNN'}  = \{x_g\mid x_g\in\text{ convex hull points labelled $y_0$} \}.\label{DB-KNN}
\end{align} 
\noindent Observe that $\repi_{g}^{kNN'}$ might contain, in the worst case scenario and particularly for high dimensional data, exactly all points of class label $y_0$ in the convex hull.
Therefore the set of representative data points can be large. Its size can be further reduced by selecting points in the decision boundary. To this end, we consider how these points spread over the decision boundary and aim to sample points that best represent the decision boundary.  
%
%

To achieve this, we consider the feature that has the largest variance, say $X_v$. Data points in $\repi_{g}^{kNN'}$ are linearly projected onto this dimension, $X_v$. For each of the $m$ equally-spaced values on $X_v$, say $\langle v_1, \ldots, v_m\rangle$, a random sampling is performed on  data points in $\repi_{g}^{kNN'}$ that have a value of $X_v$ close to $v_i$. Ultimately, a set of data points $x_m$ for each $v_i$ is selected.  This way, the representative data points to the decision boundaries will be spread over the decision boundary.
%
Alternatively, data points can be obtained by iteratively using the features ranked by variance or any other metrics (such as feature importance). This would work like a $k$-dimensional tree until a single data point can be found. In this case, no random sampling is required. 

%
Let $t_d$ be a threshold value. 
The final step is to weigh the points in $\repi_g^{kNN}$. The \freddy{rationale} is that contour points closer to the test data point $x_0$ are more useful in explaining the prediction of $x_0$. The weighting can be achieved using the distance between the points and $x_0$.
\begin{align}
\begin{split}
\label{knng} 
\repi_{g}^{kNN}  = \bigcup_{j=1}^{m} \{ & \frac{1}{1+d(x_0, x_k)}\cdot x_k ~\mid~  x_k\in\repi_{g}^{kNN'}~\wedge~ \\
&  \lvert X_v(x_k) - v_j\rvert \le t_d \},
\end{split}
 \end{align}  
where $X_v(x_k)$ denotes the value projecting $x_k$  on $X_v$.

\vspace{0.1cm}
\noindent {\bf $\bullet$ Uniform and Contrastive Explanations}. 
Computing representative data points depends on the input class label, e.g., $y_0$ in the previous descriptions. We compute not only representative points that are of the same class label $y_0$, but also compute points that are predicted with the different class label. By observation $O_2$, humans need to see more than just \emph{uniform} explanations, i.e.,  \emph{contrastive} explantions, i.e., why the input is not labeled $y_1$ and alike. For binary classification, we define the class that $x_0$ is labelled to be the positive class and the other is the negative class. The uniform and contrastive explanations are, for simplicity, called positive and negative explanations, respectively. Consequently, the representative data points with respect to $y_0$ are the positive points $\repi^{+}$. For negative class, $\repi^{-}$ can be computed analogously: we just need to change the labels from $y_0$ to $y_1$ in \eqref{DBLR1}, \eqref{locallr}, and \eqref{DB-KNN}. The two sets of data points thus serve as positive and negative evidences for explaining why $x_0$ is of class $y_0$. 

\begin{figure}[ht]
    \centering
    \begin{subfigure}[b]{0.16\textwidth}
        \includegraphics[width=\textwidth]{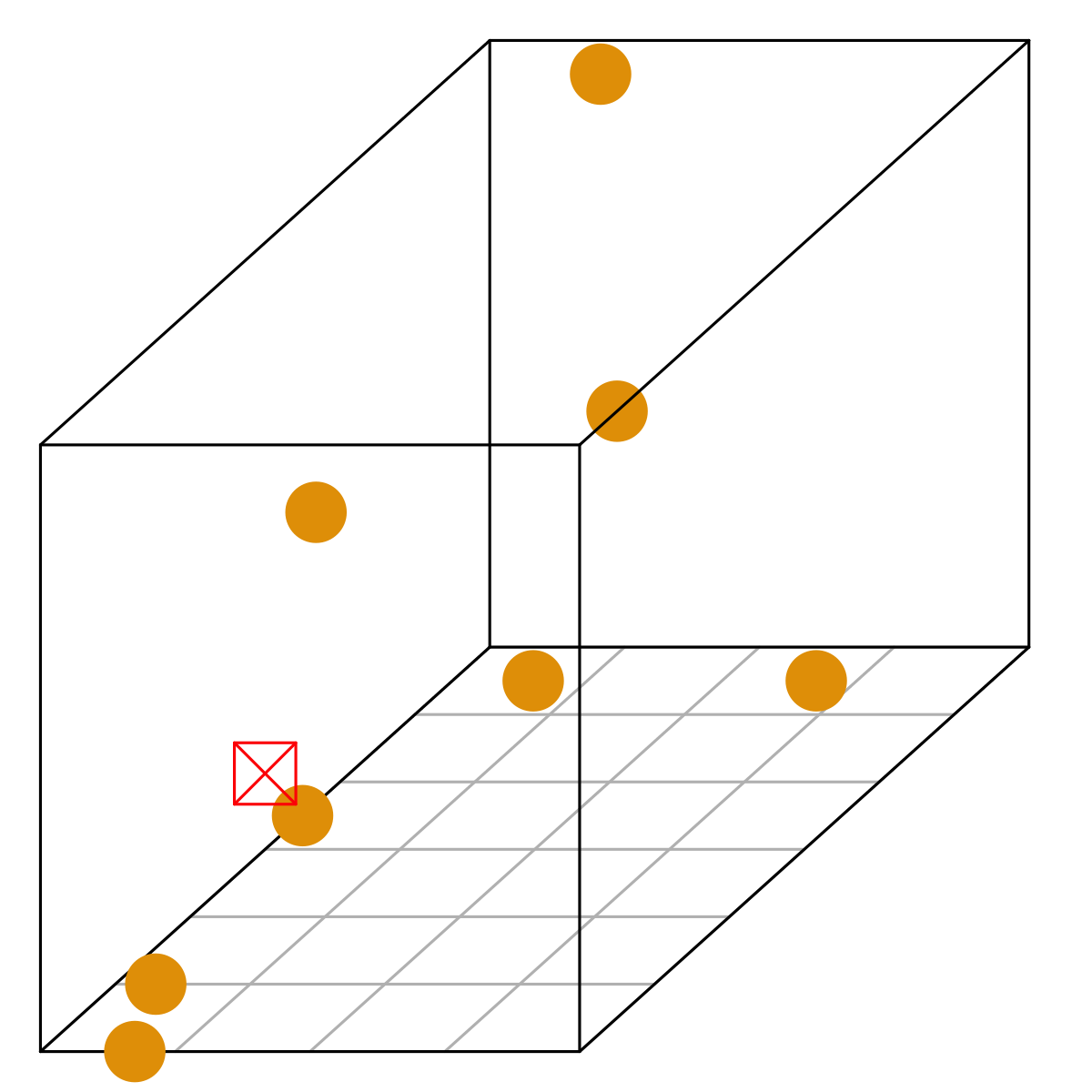}
        \caption{Extreme Positive}
        \label{fig:gp}
    \end{subfigure}
    \begin{subfigure}[b]{0.16\textwidth}
        \includegraphics[width=\textwidth]{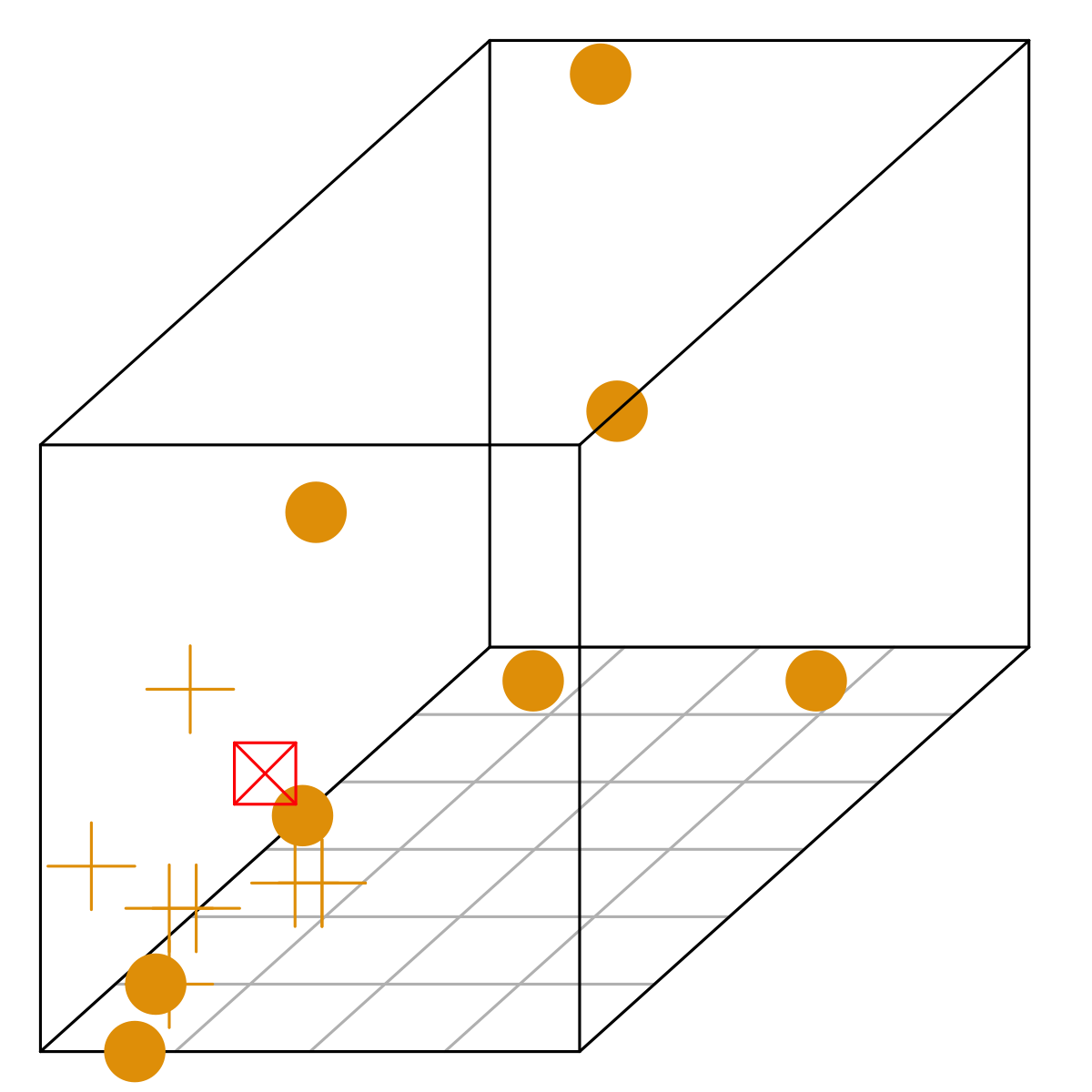}
        \caption{Local Positive}
        \label{fig:gplp}
    \end{subfigure}
\\
    \begin{subfigure}[b]{0.16\textwidth}
        \includegraphics[width=\textwidth]{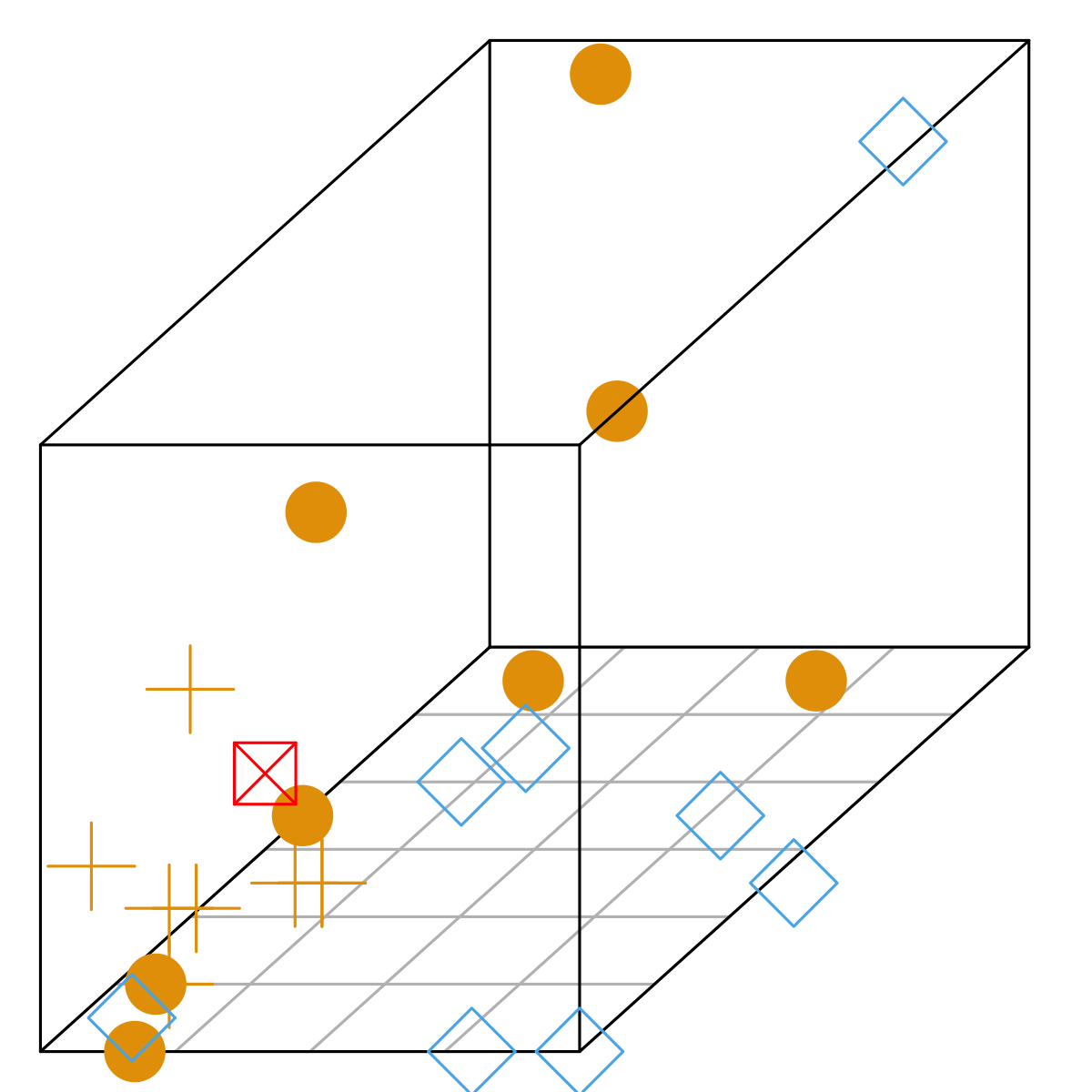}
        \caption{Extreme Negative}
        \label{fig:gplpgn}
    \end{subfigure}
    \begin{subfigure}[b]{0.16\textwidth}
        \includegraphics[width=\textwidth]{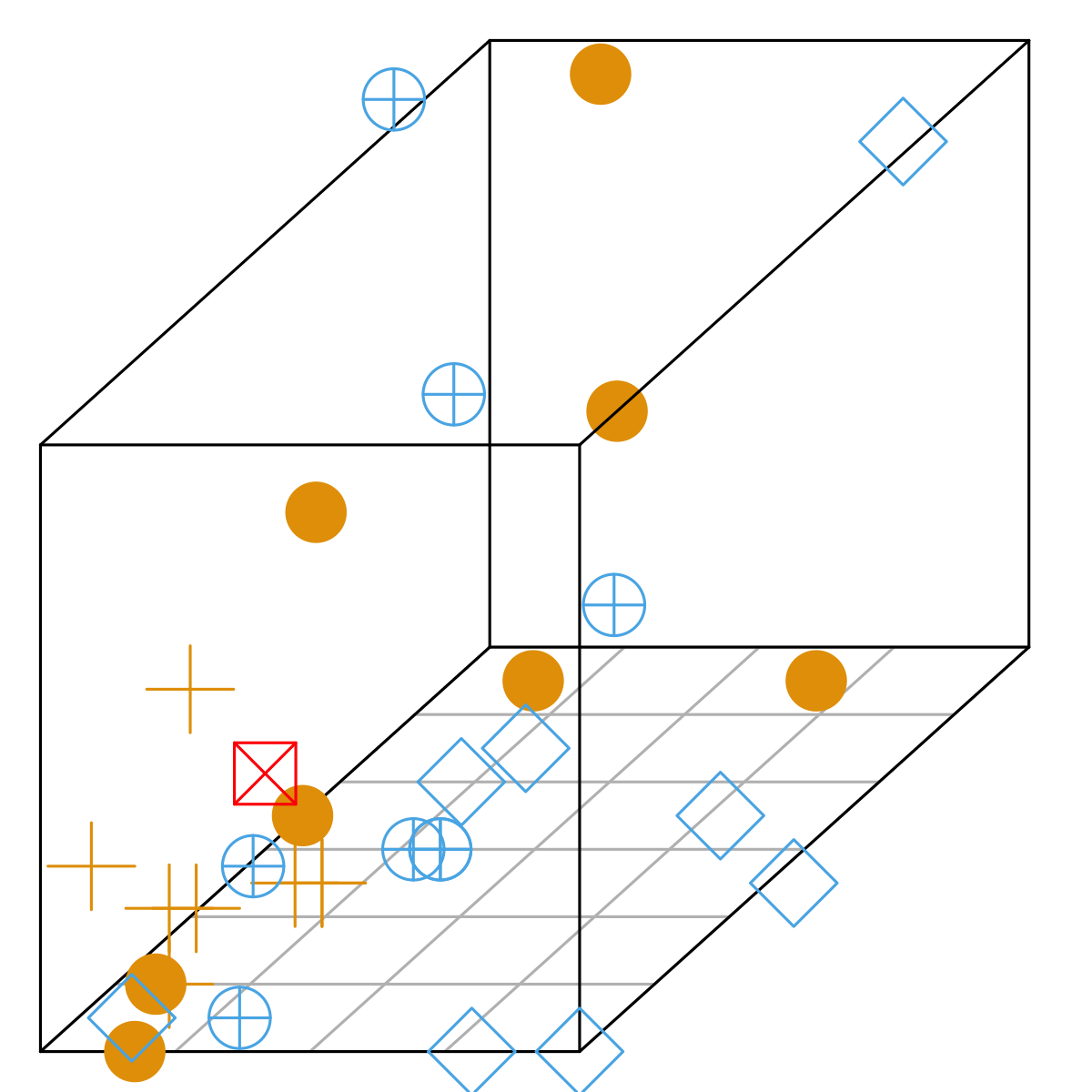}
        \caption{Local Negative}
        \label{fig:gplpgnln}
    \end{subfigure}
    \caption{k-NN Representative Data Points. The cross-squared point is the test point.}
    \label{fig:progress}
\end{figure}

\begin{ex}{\bf (k-NN Representative Data Points)}\\
Figure~\ref{fig:progress} shows the various steps of representative data points discovery.
Assume the test data point is predicted to be the positive class and no more than 8 points are to be chosen in each step.

Figure~\ref{fig:gp} first computes some points of positive labels that spread over the convex hull (decision boundary). Note that there are 42 data points in the convex hull, but only 8 points are sampled from the 42 points to approximate the convex hull, based on the spread feature, the age of patients. These 8 points are considered to be the uniform evidences, i.e., they form the set $\repi_{g}^{kNN}$ as given in \eqref{knng}.

Figure~\ref{fig:gplp} then further computes the neighboring points (positive local evidences) that are also in the positive class. These points, denoted by plus signs, form the set $\repi_{l}^{kNN}$.

After collecting all the positive evidences, Figures~\ref{fig:gplpgn} and \ref{fig:gplpgnln} show the additional data points in the negative class (contrastive information) based on the convex hull and the local neighborhood, which form \emph{negative} extreme ($\repi_{g}^{kNN-}$) and local ($\repi_{l}^{kNN-}$) evidences, respectively. 

Figure \ref{fig:gplpgnln} gives a nice visualisation of our representative selection idea: extreme  evidences are spread globally, while local evidences gather around the test data points. In addition, negative evidences appear visually more distant from the test point than positive ones.  

\end{ex}

\section{Explaining Predictions}\label{sec:Core}

%
Definition~\ref{defn:explanation} provides the basis for constructing informative explanations. To design the explanation algorithm, a few prerequisites need to be elaborated. 
In particular, the role of a domain knowledge base (or ontology) is indispensable in that semantic abstraction of data points is drawn from the knowledge base.
%

\noindent {\bf $\bullet$ Context:}
%
The explanation algorithm is formulated as $g(\mlmodel{}, X, x_i, {\ont{}}, \semuplift)$, where $\mlmodel{}$ is a classifier, $X$ is the set of training data points, $x_i$ is the input data point, $\ont$ is a {domain ontology}, {and $\semuplift$ is the semantic uplift function as given in Definition~\ref{defn:explanation}.} From $\mlmodel{}$, $X$, and $x_i$, {(\ref{DBLR1}-\ref{locallr}) and (\ref{DB-KNN}-\ref{knng})}
provide a way to compute two sets of data points (evidences) based on the decision boundaries of $\mlmodel{}$. The positive and negative evidences, denoted $\repi^{+}$ and $\repi^{-}$ respectively, can then be used to {drive the extraction of relevant information for explanation}. {We} discuss how an ontology $\ont$ can be used to abstract the semantics of these data points, which are leveraged to compute explanations. {Our approach} can be applied independently to $\repi^{+}$ and $\repi^{-}$, thus,  the general notation $\repi$ denotes either set.

\vspace{0.1cm}
\noindent {\bf $\bullet$ Notations:}
To ease the presentation, $\repi$ is given as a matrix-like structure $\matrixcps{}$ of size $m\times n$, where $\lvert\repi\rvert = m$ and there are $n$ predictors.    A single row in $\matrixcps{}$ is a data point, represented as  a set of feature-value pairs, with a weight given as $\alpha_i$ computed from \eqref{knng}. 
\begin{equation}
\alpha_i \cdot x_i:\bigwedge_{i=1\ldots n} f_i=v_i
\end{equation}

\vspace{0.1cm}
\noindent {\bf $\bullet$ Semantic Uplift of Data Points:}
There have been much prior work on deducing concepts from relational-style data. We use the existing work to uplift data semantically.
\freddy{For each $f_i=v_i$, we aim at finding its Basic-level Categorization \cite{DBLP:conf/cikm/WangWWX15}, denoted by $\text{blc}(f_i=v_i)$ with respect to a domain ontology $\ont$. Categorization is achieved in two steps. First, concepts $\text{blc}(f_i=v_i)|_{G}$ are identified in a large knowledge graph $G$ i.e., a graph dominated by instance and \emph{is-a} relationships such as Dbpedia \cite{DBLP:journals/semweb/LehmannIJJKMHMK15} and Microsoft Concept Graph \cite{DBLP:conf/sigmod/WuLWZ12}
following \cite{DBLP:conf/ijcai/WangZWMW15}.
Then a mapping step from concepts $\text{blc}(f_i=v_i)|_{G}$ to $\cp{i}$ in the domain ontology $\ont$ is required to contextualized categorization in a targeted domain.
In other words a concept $\cp{i}\in\ont$ is identified such that:
\begin{equation}\label{eq:mapping}
\cp{i} \doteq m(\text{blc}(f_i=v_i)|_{G})
\end{equation}
where $m$ is a mapping function from $G$ to domain ontology $\mathcal{O}$. 
The concept mapping is achieved following \cite{ehrig2004qom} where both syntactic and semantic similarities (distance among similar concepts) are considered.
Therefore the semantic uplift function $\semuplift$ in Definition \ref{defn:explanation} is computed through the composition of $m$ and blc i.e., $\semuplift \doteq m \circ \text{blc}$.
We adopted a \freddy{2-step} process to ensure a maximum coverage of $f_i=v_i$ in $\ont$. Indeed a more direct approach from $f_i=v_i$ in $\ont$ could result in no mapping, and then no semantic association for $f_i=v_i$. 
The knowledge graph layer provides a much larger \freddy{input set} to be mapped in $\ont$, and then a better semantic coverage for $f_i=v_i$.
}

\vspace{0.1cm}
\noindent {\bf $\bullet$ Representation of Data Points:}
\noindent Since not all $f_i=v_i$ can be automatically matched with concepts in $\ont$, we differentiate the semantic and non-semantic parts. To this end we assume there is a set $K\subseteq\{1, \ldots, n\}$ such that the final representation of a data point can be defined as two components, which are the projections of points for the  feature-value pairs that cannot be matched with ontological concepts ($\prj{f}{x_i}$) and the semantic counterpart ($\prj{c}{x_i}$), respectively.
\begin{align}{\label{eq:RDPs}}
\begin{split}
x_i: &\bigwedge_{k\in K}\alpha_i\cdot\{f_{ik}=v_{ik}\}\wedge\bigwedge_{j\in\{1, \ldots, n\}\backslash K}\alpha_i\{\cp{ij}\} \\
 & = \prj{f}{x_i}\wedge\prj{c}{x_i}
 \end{split}
\end{align}
Projections are applied to each row in $\matrixcps{}$, and the concept components of all rows forms a set, $\setcps{in}$, that serves as the input of our explanation approach (cf. Algorithm \ref{alg:core}).
\begin{equation}\label{eq:inputconcepts}
\setcps{in} = \bigcup_{i\in[1, m]}\prj{c}{x_i}
\end{equation}

Note that for duplicate concepts across data points, the weights of these concepts will increase accordingly. For positive data points $\repi^{+}$, the set is $\setcps{in}^{+}$. Similarly, $\setcps{in}^{-}$ is computed for negative data points $\repi^{-}$.

\begin{ex}{\bf (Semantic Uplift)}\\
\freddy{
The features of Haberman's survival data set have a natural categorization of values using wikipedia as $G$ in \eqref{eq:mapping}. For instance }
$TheSilentGeneration$ denotes people born between $1925$ and $1941$. Patient $16$, \freddy{classified as a patient who survived 5 years or longer,} is described as \eqref{eq:p21}, and as \eqref{eq:p22} after semantic encoding of the data point. \freddy{Note that the study has been conducted between $1958$ and $1970$ hence a $35$ year-old patient in $1963$.}
\begin{align}
p_{16} :~ & age = 35 \sqcap yearOp=1963\nonumber\\
            & \sqcap  numberNodes = 0\label{eq:p21}\\
p_{16} :~ & TheSilentGeneration\sqcap OperationIn1960s\nonumber\\
            & \sqcap NoPosAxillaryNode\label{eq:p22}
\end{align}

Semantic representations of human population have been extracted\footnote{https://en.wikipedia.org/wiki/Generation\#List\_of\_generations} for appropriate semantic mapping.
\end{ex}

\vspace{0.1cm}
\noindent {\bf $\bullet$ Explanation Concept Completion} The input concepts given in (\ref{eq:inputconcepts})  are not necessarily easy to understand for two reasons: \emph{a}) they tend to be loosely connected to each other as not all feature value pairs can be semantically uplifted. \emph{b}) these concepts may be data specific due to the semantic uplift so that humans may not understand the low-level concepts well. To address these issues, we show how to introduce more human-comprehensible and semantically connected concepts from the ontology. Furthermore, to optimize the completion process and to ensure the final explanation concepts are succinct,  the following constraints are stipulated:
%
\begin{itemize}[leftmargin=*,labelindent=1em,labelsep=1em]
\item[$C_1$] Minimize the size of $e$ for succinct explanations.
\item[$C_2$] Maximize the number of matching among input concepts and ontological concepts.  
\item[$C_3$] Maximize the total weight of matching. 
\end{itemize}

Note that the concepts in the input  $\setcps{in}=\{\alpha_i \cp{i}\}$ are weighted. To fully leverage the semantics of concepts, the structure of $\ont$ is used to find concepts that can abstract input concepts. Our graph-based traversal requires the following notions for defining relationships among any concepts.  

\begin{defn}{\bf (Distance between Concepts)}{\label{defn:DC}}\\
Suppose $\ont$ be an ontology represented as a graph.
Given two weighted concepts $\alpha_1\cp{1}, \alpha_2\cp{2}$ over graph $\ont$. The distance $\dist{\cp{1}}{\cp{2}}$ between $\cp{1}$ and $\cp{2}$ is defined to be the minimum length of the path from $\cp{1}$ to $\cp{2}$. 
\end{defn}

\begin{defn}{\bf (Concept Matching)}{\label{defn:SPBC}}\\
Let a mapping be a partial function $\conceptmapping:\setcps{in} \rightarrow \setcps{out}$, which defines a set of matching between concepts in $\ont$. A matching of $(\cp{1}, \cp{2})$ is the shortest path {following any labelled edges} in $\ont$ from $\cp{1}$ to $\cp{2}$.
\end{defn}

Note that each edge on the path of a matching carries a  weight, $\lambda_i$, that denotes the semantic relatedness between concepts. For a matching $(\cp{1}, \cp{2})$ of distance one, an aggregated weight of $\cp{2}$ can be computed as follows, e.g., $\gamma_i = \sum_{1}^{k}\beta_k\cdot\lambda_k$   for a concept with matching from $k$ concepts in $\ont{}$, each of which has a weight $\beta_i$. In the initial case, i.e., concepts in $\setcps{in}$, the weight $\beta_i = \alpha_i$.  

%
Following Definitions \ref{defn:DC} and \ref{defn:SPBC}, we are ready to instantiate the constraints $C_1$-$C_3$. Let $\dist{\cp{1}}{\cp{2}}$ denote the number of hops between two nodes in $\ont$ and $\gamma_i$ denote the weight of $\cp{i}$. The algorithm aims to find a set of output concepts  defined as follows:
\begin{align}\label{optimalConditions}
\begin{split}
&~~~~~~~~~~~~~~\setcps{O}=\argmax_{\setcps{e}}(a_1\score{v} + a_2\score{l}+a_3\score{d}),\\
&\text{where  $a_{i, i\in\{1, 2, 3\}}$ are weighting factors, and } \\
&\score{v} = \frac{1}{\lvert \conceptmapping^{-1}(\setcps{e})\rvert}, 
~~\score{l} = \frac{1}{\lvert\setcps{e} \rvert}, \\
 &\score{d} =  \sum\frac{\gamma_i}{\dist{\cp{j}}{\cp{i}}} \text{ for any $\conceptmapping(\cp{j}) = \cp{i}$.}
\end{split}
\end{align}
For tractable computation, our algorithm uses random hill climbing to find  $\setcps{O}$, as discussed in the next section.
%
 

%

%
\subsection{Computing Informative Explanations}\label{sec:ExplanationAlgorithm}

The main algorithm, Algorithm~\ref{alg:core}, computes the explanation concept completion based on (\ref{optimalConditions}). 

\vspace{0.1cm}
\noindent {\bf $\bullet$ Input:}
The input to the algorithm \freddy{includes} a set of concepts $\setcps{I}$, an ontology $\ont$ represented as a graph, a given integer $k$ to restrict the depth for traversing $\ont$, and two additional control parameters, $h$ and $mp$. 
\begin{algorithm}[h!] 
\caption{The Algorithm for Explanation Concept Completion. We define $\score{total}$ to be the function $a_1\score{v} + a_2\score{l}+a_3\score{d}$ in (\ref{optimalConditions}). \label{alg:core}}
\SetAlgoLined
\SetKwInOut{Input}{Input}\SetKwInOut{Output}{Output}
\Input{$\setcps{I}$, $\ont$, $k$, $h$, $mp$}
\Output{$\setcps{O}$}
$\setcps{I} \leftarrow removeDuplicate(\setcps{I})$\;
$V \leftarrow sort(\setcps{I})$\; 
\For{$i=1$ to $\lvert V\rvert$}
{ 
\If{there is a matching $(\cp{i},\cp{i+1})$ in $\ont$}
{ $V \leftarrow V\backslash\cp{i}$ }
}
$s_{p} = 0,\;\;\setcps{O} =  \emptyset$\;
\For{$i=1$ to $h$}
{ 
$V_{i}\leftarrow randomSubset(V)$\;
$lv=1, V_{p} = V_{i}$\;
\While{$V_i'\neq\emptyset$ and $lv\le k$}
{
$V_i' \leftarrow \{\cp{b} \mid \exists (\cp{1},\cp{b}), (\cp{2},\cp{b})$ of distance 1 and $\{\cp{1},\cp{2}\}\subseteq V_p\}$\;
$V_i' \leftarrow first(sort(V_i'), mp)$\; 
$lv=lv+1$\;
\lIf{$V_i'\neq\emptyset$}{$V_p = V_i'$}
}
\lIf{$\score{total}(V_p) > s_p$}{$ s_p= \score{total}(V_p), \setcps{O} = V_p$}
}
\end{algorithm}


\vspace{0.1cm}
\noindent {\bf $\bullet$ Algorithm~\ref{alg:core}:}
Line 2 sorts the concepts decreasingly by weights. Lines 3-7 remove concepts that are subsumed by any concepts in $V$ because the purpose of our algorithm is to uplift special concepts into more general ones. The loop in lines 9-19 is a random-restart step to reduce the exponential search space of subsets of $V$ to $h$ restarts. Here, $h$ specifies the number of restarts desired. Line 10 obtains a random subset of $V$, which is then used to find matching successors in $\ont$ as shown in lines 12-17. Line 13 collects matching concepts that can match at least two different concepts so as to further limit the search space. There is also an implicit condition to ensure correctness: $V_i'$ should collect only concepts that have never been collected before in each restart, due to possible cycles in the ontological graph. 
Line 14 first sorts the matching concepts and then \freddy{picks} the first $mp$ concepts. Here $mp$ specifies the number of concepts to be chosen for next matching. This step is necessary as potentially all concepts in $\ont$ can be in $V_i'$, so a constant $mp$ can significantly reduce the search space. The output $\setcps{out}$ is the set of matching concepts that maximizes the weight.

\vspace{0.1cm}
\noindent {\bf $\bullet$ Contraction}
Applying Algorithm~\ref{alg:core} to $\setcps{in}^{+}$ results in $\setcps{out}^{+}$. 
For $\setcps{in} = \setcps{in}^{+}\cup\setcps{in}^{-}$, we also apply Algorithm~\ref{alg:core} to $\setcps{in}^{-}$ and obtain $\setcps{out}^{-}$. 
The two sets of concepts form the basis to provide both uniform and contrastive explanations, each set being a \emph{group} of explanations in that class. Note that in case of multi-classification, there will be many groups of contrastive explanations. To avoid excessive contrastive evidences, it is reasonable to restrict the groups of contrastive explanation to one or two. This can be realized by selecting the next one or two most probable class labels predicted by the classifiers.

Now consider the binary classification case. The uniform explanations may contain knowledge already entailed by the contrastive explanations. It makes sense to keep only the essential information in the uniform explanations for succinctness. As an example, assume the uniform explanations have only one concept $\{GraduateStudent\}$ and the contrastive explanations have $\{{PhDStudent}\}$, a more informative uniform explanation would be \emph{``$GraduateStudent$ but not $PhDStudent$"}, which means $\{MasterStudent\}$ w.r.t. a common sense ontology.

For this purpose \emph{concept difference} \cite{dlhb} is used to find out the concepts that entail some positive concept but not any of the negative concepts. 
Given concepts $\cp{p},\cp{n}$ from  $\setcps{out}^{+},\setcps{out}^{-}$, respectively, the difference between two concepts is computed as follows:
\begin{align}\cp{p}\backslash\cp{n} =   \bigcup\{\cp{d} ~\mid~  \cp{d}\sqsubseteq\cp{p}\text{ and } \cp{d}\not\sqsubseteq\cp{n} 
\}  \label{eq:defdiff}
\end{align}
The subsumption relation $\sqsubseteq$ may introduce  many unseen ontological concepts as difference concepts. To select the useful difference concepts, these concepts are ranked according to a weight that indicates how closely a difference concept is semantically related to concepts found in the data. We define the importance of each $\cp{d}$ to be:
\begin{align}\label{eq:importance}
imp(\cp{d}) = \frac{1}{ n} \sum_{v=1}^{n} \alpha_v\dist{\cp{d}}{\cp{v}} \text{ for } \cp{v}\in\setcps{out}^{+},
\end{align} 
\noindent where $n$ is the size of the input $\setcps{out}^{+}$.
The final set of difference concepts used to replace the grop $\setcps{out}^{+}$, with a defined threshold $\delta$, is:
\begin{align}\label{eq:diffrep}
\begin{split}
\setcps{diff}^{+} =  \bigcup\{\cp{p}\backslash\cp{n}\mid &~\cp{p}\in\setcps{out}^{+}, \cp{n}\in\setcps{out}^{-},\\
& \text{ and } imp(\cp{p}\backslash\cp{n}) \ge\delta \}
\end{split}
\end{align}
%

\noindent {\bf $\bullet$ Ranking} The grouping step generates two groups, $\setcps{diff}^{+}$ and $\setcps{out}^{-}$. Recall that humans select explanations \emph{relevant} to them. To allow for human subjectivity, we associate all concepts with a rank order within the same group such that concepts across different groups with the same rank order represent a single rational explanation. We show one possible method to compute the rank orders: 
\begin{enumerate}
\item A dense rank is computed for concepts in $\setcps{diff}^{+}$ based on the importance defined in (\ref{eq:importance}).
\item For any $\cp{n}\in\setcps{out}^{-}$ used in computing $\setcps{diff}^{+}$, i.e.,  (\ref{eq:diffrep}), the rank order of $\cp{n}$ is the same as that of the majority of the results $\cp{d}\in\setcps{diff}^{+}$. Note a single $\cp{n}$ may be used to compute many difference concepts. 
\item For any $\cp{n}\in\setcps{out}^{-}$ \emph{not} used in (\ref{eq:diffrep}), the rank order of $\cp{n}$ is the same as the majority order of the most similar difference concepts. Note that an upper limit threshold, $\sigma$, is set to ensure concepts are sufficiently similar to some difference concepts. Otherwise, next step is required.
\begin{align*}
\{\cp{d}~\mid~&\cp{d}\in\setcps{diff}^{+}, \dist{\cp{n}}{\cp{d}}<\sigma, \neg\exists \cp{d'}\in\setcps{diff}^{+}  \\
& \text{ s.t. } \dist{\cp{n}}{\cp{d'}} < \dist{\cp{n}}{\cp{d}}\}.
\end{align*}

\item Otherwise, the rank order of any $\cp{n}\in\setcps{out}^{-}$ is the next order to the lowest order in $\setcps{diff}^{+}$.
\end{enumerate}

After ranking, concepts of the same order from both the uniform and contrastive explanations form an explanation of that rank order. Human users can choose, among these succinct, informative explanations, the ones that they believe to be most relevant using the rank order.

\begin{ex}{\bf (Informative Explanations)}\\
Consider the test data point given in \eqref{eq:exFinal}, representing a $30$ year-old individual with an operation which occurred in $1964$, and a number of nodes equal to $1$. %
\begin{align}{\label{eq:exFinal}}
p_1: ~& age=30 \sqcap yearOp=1964 \nonumber\\
         & \sqcap numberNodes = 1
\end{align}
Although concepts are not drawn from the test point, it is easy to see that $p_1$ is also in ``TheSilentGeneration", had ``OperationIn1960s," and had ``OnePosAxillaryNode."

The objective is to understand why such individual is classified to be positive, i.e., survived 5 years or longer.
The concepts from the representative data points fall into  positive \eqref{eq:plus} and negative \eqref{eq:minus} groups. Computing the concept difference in this example discards the concept ``NoPosAxillaryNode", though it does not introduce new concepts \eqref{eq:expdiff}:
{\footnotesize
\begin{align}
\setcps{out}^{+} & = \{0.9TheSilentGeneration,
0.7OperationIn1960s,\nonumber\\
                 & 0.5NoPosAxillaryNode\}\label{eq:plus}\\
\setcps{out}^{-}& = \{ 0.6TheGIGeneration, 0.3OperationIn1950s,\nonumber\\ 
                 &  0.5OperationIn1960s, 0.5NoPosAxillaryNode\}\label{eq:minus}\\
 \setcps{diff}^{+} & = \{0.85TheSilentGeneration
\}\label{eq:expdiff}         
\end{align} 
}
The weights are computed by semantic uplifts and also reflect the proportion of data points that share the concepts cf. \freddy{\eqref{eq:importance}}, e.g., in the positive data points there are more patients in the silent generation, while the G. I. generation is the majority in the negative data points.

The final explanations were ranked within groups:
\vspace{0.5em}\\
\begin{tabular}{ccc}
 & $\setcps{diff}^{+} $ & $\setcps{out}^{-} $ \\
\noalign{\smallskip} \hline\noalign{\smallskip}
 rank$_1$ & $TheSilentGeneration$ & $TheGIGeneration$\\
 \noalign{\smallskip} \hline\noalign{\smallskip}
\multirow{3}{*}{rank$_2$}&  & $OperationIn1960s$ \\
&  & $OperationIn1950s$ \\
&  & $NoPosAxillaryNode$ \\
\end{tabular}
\vspace{0.5em}\\
The rank orders of concepts in the contrastive explanations $\setcps{out}^{-} $ are derived from their semantic similarity to the only uniform explanation $TheSilentGeneration$, as given in Steps 3\&4 of the ranking process.
\end{ex}  
{\section{Conclusions and Future Work}}\label{sec:ConclusionFutureWork}

Our approach, exploiting the semantics of data points, tackles the problem of explaining the predictions in an informative manner to human users. 
Semantic reasoning and machine learning have been combined
by revisiting decisions boundaries and its representative elements  as semantic characteristics of explanations. Such characteristics are then leveraged to derive informative explanations with respect to the domain ontologies. The core contributions of our approach include its ability to capture interesting data points that exhibit extremity (in the form of decision boundaries) and local context of the test data points (in the form of neighborhood) and its manipulation of semantics to enhance  \emph{informativeness} for human-centric explanations.

To generalize our approach for multi-label classifiers, the key difference is on the computation of $\repi^+$ and $\repi^-$. For a particular test point, all the predicted labels are considered to be positive classes.  Computing $\repi^+$ is equivalent to computing the union of $\repi^+$ in each positive class. For computing $\repi^-$,  it can be done for each negative class \emph{separately}. So the contrastive evidences are for all positive labels against each and every of the negative labels.  However, we advise choosing only the top few negative classes (by user specification or some predefined popularity metrics of classes).

There are several extensions to be considered in our future work. First, explanation relevancy can be improved by considering user profiles, instead of allowing for user choices. Second, we will investigate other types of machine learning models, for instance, random forest classifiers.

\newpage
\bibliographystyle{aaai}
\bibliography{exp18}

\end{document}